\documentclass[10pt,conference,letterpaper]{IEEEtran}
\IEEEoverridecommandlockouts
\usepackage{times,amsmath,epsfig}
\usepackage{bm, amsmath,amssymb,graphicx}
\usepackage{subfigure}
\usepackage{algorithm, algpseudocode, algorithmicx}
\usepackage{multirow}
\title{Smoothness-Constrained Image Recovery from Block-Based Random Projections}
\author{%
{Giulio Coluccia, Diego Valsesia, Enrico Magli\thanks{This research activity has received funding from the European Research Council under the European Community's Seventh Framework Programme (FP7/2007-2013) / ERC Grant agreement n° 279848\vspace*{-3mm}}}%
\vspace{1.6mm}\\
\fontsize{10}{10}\selectfont\itshape
Dipartimento di Elettronica e Telecomunicazioni\\
Politecnico di Torino, Italy\\
\fontsize{9}{9}\selectfont\ttfamily\upshape
}

\newcommand{\mat}[1]{\ensuremath{\bm{\mathrm{#1}}}}
\newcommand{\A}{\ensuremath{\mat{A}}}

\newcommand{\x}{\ensuremath{\mat{x}}}
\newcommand{\X}{\ensuremath{\mat{X}}}

\newcommand{\y}{\ensuremath{\mat{y}}}

\newcommand{\Ps}{\ensuremath{\bm{\Psi}}}

\newcommand{\Ph}{\ensuremath{\mat{\Phi}}}

\newcommand{\trasp}[1]{\ensuremath{#1 ^\mathsf{T}}}

\newcommand{\vect}[1]{\ensuremath{\mathrm{vec}\left \{ #1\right\}}}

\newcommand{\N}{\mathcal{N}}

\def\Ri{\mathbb{R}}

\newcommand{\lzeronorm}[1]{\ensuremath{\left\| #1\right\|_{\ell_0}}}
\newcommand{\lonenorm}[1]{\ensuremath{\left\| #1\right\|_{\ell_1}}}
\newcommand{\ltwonorm}[1]{\ensuremath{\left\| #1\right\|_{\ell_2}}}

\begin{document}
\maketitle

\begin{figure}[b]
\parbox{\hsize}{\em
MMSP'13, Sept. 30 - Oct. 2, 2013, Pula (Sardinia), Italy.

978-1-4799-0125-8/13/\$31.00 \ \copyright 2013 IEEE.
}\end{figure}

\begin{abstract}
In this paper we address the problem of visual quality of images reconstructed from block-wise random projections. Independent reconstruction of the blocks can severely affect visual quality, by displaying artifacts along block borders. We propose a method to enforce smoothness across block borders by modifying the sensing and reconstruction process so as to employ partially overlapping blocks. The proposed algorithm accomplishes this by computing a fast preview from the blocks, whose purpose is twofold. On one hand, it allows to enforce a set of constraints to drive the reconstruction algorithm towards a smooth solution, imposing the similarity of block borders. On the other hand, the preview is used as a predictor of the entire block, allowing to recover the prediction error, only. The quality improvement over the result of independent reconstruction can be easily assessed both visually and in terms of PSNR and SSIM index.
\end{abstract}

\section{Introduction}

Conventional image acquisition and compression schemes usually rely on the sampling of a huge number of pixels satisfying the classic Shannon/Nyquist theorem. Then, the size of the acquisition is reduced by the means of a more energy--compacting signal representation, usually consisting in the projection of blocks of pixels onto a convenient basis, like the DCT or the wavelet. Recently, the theory of Compressed Sensing (CS) \cite{candes2006nos, donoho2006cs} has proposed a new paradigm. According to this theory, a sparse or compressible signal (as natural images are), can be sensed acquiring a small number of random projections, called \emph{measurements}, and recovered from these measurements using algorithms promoting sparsity. While on one hand the acquisition process is really simple, the complexity is transferred on the decoder, where the optimum reconstruction algorithm, based on $\ell_0$ norm minimization, has combinatorial complexity, while its $\ell_1$ norm approximation has a complexity which is roughly cubic with the size of the signal. Indeed, this or even computationally simpler algorithms, like OMP \cite{omp}, are far too expensive to deal with the reconstruction of a signal whose dimension is large, as is the case of images or, more generally, multidimensional signals.

Different approaches have been proposed when applying CS to multidimensional signals, based on the partitioning of the signal to be acquired, in order to reduce the size of the signal to be reconstructed. For example, in \cite{coluccia2012progressive} a raster scanning approach has been combined to a reconstruction scheme based on linear predictors. 

A different approach to the problem is to partition the image in blocks, as is done in the JPEG or MPEG standards for image and video encoding, respectively, where an orthonormal transform is applied to non overlapping blocks of pixels of the image or of a frame. The same approach has been proposed for compressed image sensing, where it was labelled as Block Compressed Sensing (BCS) after \cite{gan2007block}. In that paper, each block of pixels is acquired using the same sensing matrix. At the decoder, each block is processed independently. This approach usually leads to poor visual quality due to blocking artifacts, especially when the number of measurements per block is low. To overcome this problem, the authors of \cite{gan2007block} propose to apply to the reconstructed image two additional processing stages, where the entire image undergoes two sequential iterative algorithms based on Projections onto Convex Sets (POCS) and Iterative Hard Thresholding (IHT). This approach was further improved in \cite{mun2009block}, where directional transforms were used instead of conventional DCT or DWT, along with an improved thresholding criterion.

In this paper, we propose a different approach to the reconstruction process. In particular, we propose to improve the reconstruction by imposing smoothness constraints between adjacent blocks, in addition to sparsity constraints for each block, in order to ``drive'' the reconstruction process towards solutions promoting both  smoothness and sparsity at the same time. This result is obtained by the efficient computation of a preview of each block, whose purpose is twofold. First, it supplies an estimation of block borders, used to impose additional smoothness constraints. Second, it is used as a predictor of the block, allowing to reconstruct prediction error, only. We dub our algorithm Smoothness-Constrained Block Compressed Sensing (SC-BCS). Simulation results show a significant performance improvement with respect to independent block reconstruction, in terms of PSNR, structural similarity (SSIM) index \cite{wang2004image} and visual quality assessment, demonstrating the validity of this approach. Compared to the results obtained in \cite{mun2009block}, our algorithm obtains better results especially when the number of measurements is low. Moreover, with respect to \cite{gan2007block} and \cite{mun2009block}, both the acquisition and the reconstruction stages can be parallelized, since each block of the image is acquired and reconstructed independently.

\section{Background}\label{sec:background}

\subsection{Notation and definitions}\label{sec:notation}

We denote (column-) vectors and matrices by lowercase and uppercase boldface
characters, respectively. The $(m,n)$-th element of a matrix $\A$ is $(\A)_{m,n}$. 
The transpose of a matrix $\A$ is $\trasp{\A}$. The stack operator $\vect{\A}$ denotes the column vector obtained by stacking the columns of \A\ on top of each other, from left to right.

The notations $\lzeronorm{\mat{v}}$, $\lonenorm{\mat{v}}$, $\ltwonorm{\mat{v}}$ denote the number of nonzero elements, the $\ell_1$-norm and the Euclidean norm of vector $\mat{v}$, respectively. The notation $a\sim\N(\mu,\sigma^2)$ denotes a Gaussian random variable $a$ with mean $\mu$ and variance $\sigma^2$~.

\subsection{Compressed Sensing}\label{sec:CS}

In the standard CS framework, introduced in \cite{candes2006nos}, a signal $\x\in\Ri^{N\times 1}$
 which has  a sparse representation in some basis $\Ps\in\Ri^{N\times N}$, \textit{i.e},
$
\x = \Ps \bm{\theta},\quad \lzeronorm{\bm{\theta}} = K,\quad K\ll N
$,
can be recovered by a smaller vector $\y\in\Ri^{M\times 1}$, $K<M<N$, of linear measurements $\y = \Ph\x$, where $\Ph\in\Ri^{M\times N}$ is the \emph{sensing matrix}. The optimum solution, seeking the sparsest vector compliant with \y\, is an NP-hard problem, but one can resort to a linear programming reconstruction by minimizing
the $\ell_1$ norm
\begin{equation}\label{eq:CS_recovery}
\widehat{\bm{\theta}}=\arg\min_{\bm{\theta}}\lonenorm{\bm{\theta}}\ \quad \text{s.t.}\quad \Ph\Ps\bm{\theta} = \y~,
\end{equation}
provided that $M\sim K\log(N/K)$.

The same algorithm can be used to reconstruct signals which are not exactly sparse, but rather compressible, meaning that the magnitude of their sorted coefficients (in some basis \Ps) decays at least exponentially.

It has been shown in \cite{baraniuk2008spr} that extracting the elements of \Ph\ at random from any sub-Gaussian distribution, allows a correct reconstruction with overwhelming probability.

An alternative approach to the $\ell_1$ norm minimization is the minimization of the Total Variation (TV) norm of the image. The TV norm of a bidimensional signal, in its isotropic version, can be defined as

\begin{equation}
\mathrm{TV}(\X) = \sum_{i,j}\sqrt{|(\X)_{i+1,j}-(\X)_{i,j}|^2+|(\X)_{i,j+1}-(\X)_{i,j}|^2}
\end{equation}

Seeking to minimize the TV norm relies on the assumption that the gradient of the image is approximately sparse, hence the TV norm should be small.

Hence, the reconstruction problem \eqref{eq:CS_recovery} becomes
\begin{equation}\label{eq:CS_TV_recovery}
\widehat{\X}=\arg\min_{\X}\mathrm{TV}(\X)\ \quad \text{s.t.}\quad \Ph\cdot\vect{\X} = \y~,
\end{equation}

\section{Proposed Algorithm}

\begin{figure}
\centering
\includegraphics[width=0.7\columnwidth]{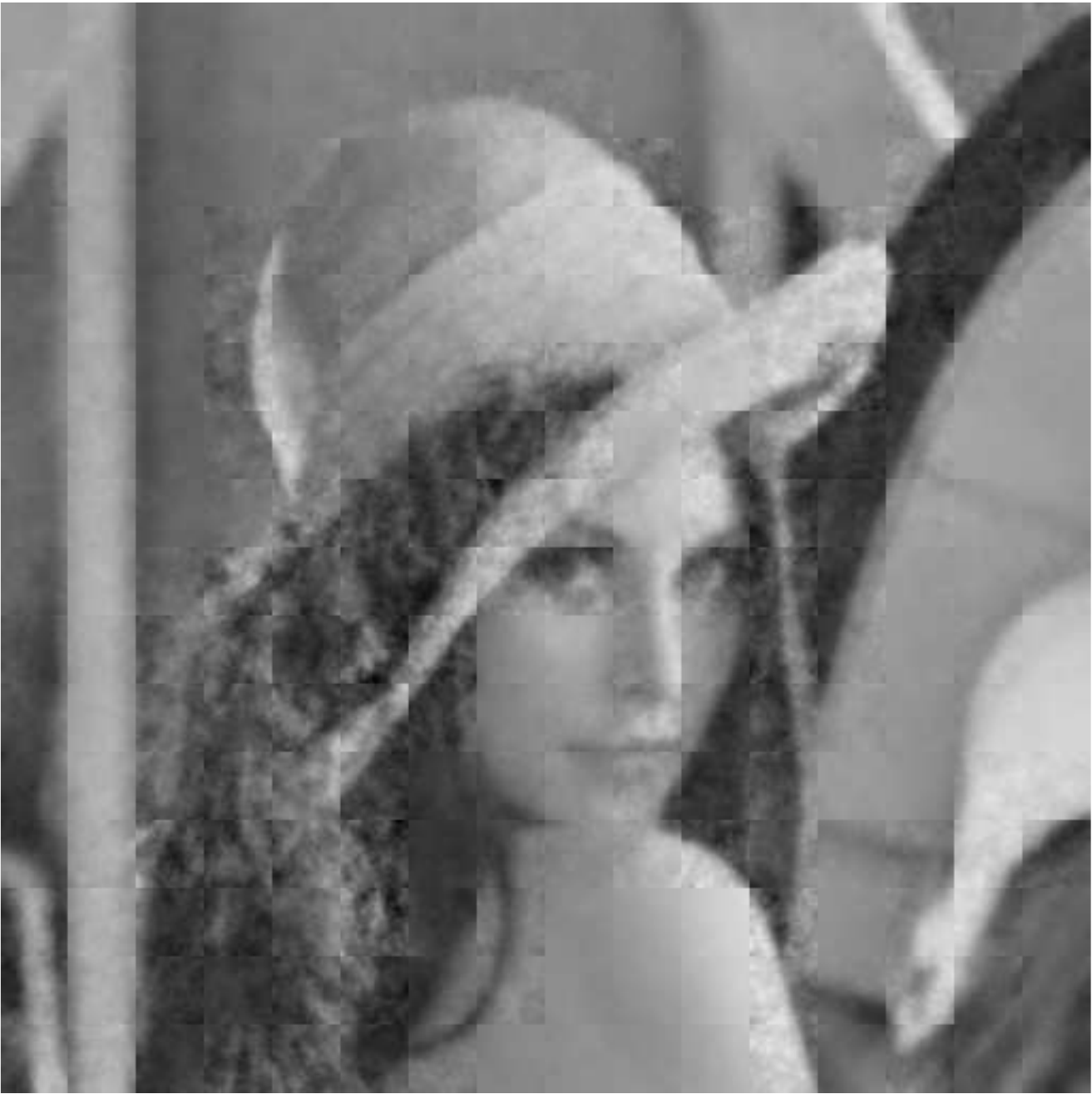}
\caption{Standard independent blockwise reconstruction, $M=64$ measurements per block}
\vspace*{-4mm}
\label{fig:standard_gauss_M64}
\end{figure}

Traditional BCS techniques rely on independent blockwise image acquisition and independent blockwise reconstruction. This may result in discontinuities in the reconstruction of adjacent blocks, especially when the number $M$ of measurements acquired for each block is low. See for example Fig.~\ref{fig:standard_gauss_M64}, where, for each $32\times 32$ block, $M=64$ measurements have been taken. The blocking artifacts have a large impact on the visual quality of the reconstruction. To overcome this problem, we propose a technique, detailed in the following sections, based on additional constraints to the reconstruction problem. The idea is to enforce smoothness between the reconstruction of block borders\footnote{Here and in the following, with the term \emph{block borders} we refer to pixels belonging to the first and last rows and to the first and last columns of the block.} and the contiguous borders of adjacent blocks, in order to reduce blocky artifacts in the reconstruction of adjacent blocks and obtain homogeneity overall. We dub our algorithm as Smoothness-Constrained Block Compressed Sensing (SC-BCS).

\subsection{Image acquisition}\label{sec:image_acquisition}

\begin{figure}
\centering
\vspace*{-5mm}
\includegraphics[width=0.84\columnwidth]{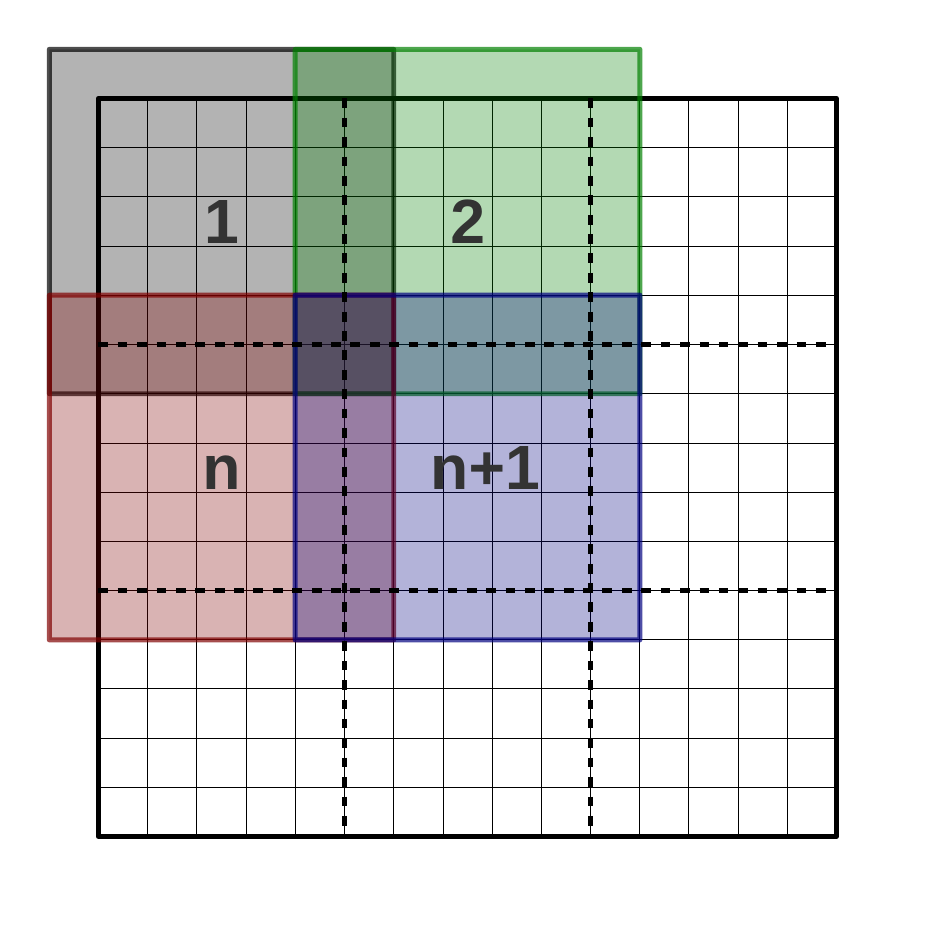}
\vspace*{-8mm}
\caption{Overlapping blockwise image acquisition}
\vspace*{-4.5mm}
\label{fig:image_acquisition}
\end{figure}

As we will explain in section~\ref{sec:image_reconstruction}, our idea relies on an initial estimation of block borders. We propose the acquisition of the image in an overlapping block fashion, in order to have an initial estimate of satisfactory quality. For each non overlapping block, we consider 1 extra pixel per side, as depicted in Fig.~\ref{fig:image_acquisition}, resulting in a global overlapping of 2 pixels per side. Each block $\X_i$ has size $B \times B$. It is raster scanned then measured with a sensing matrix \Ph\ as
$
\y_i = \Ph_i \cdot\vect{\trasp{\X_i}}~,
$
where $\y_i$ is the vector collecting the $M$ measurements of block $\X_i$ and $\Ph_i$ is the sensing matrix of block $\X_i$. As for the sensing matrix, we choose a matrix allowing us to obtain an estimate of the block borders before performing CS reconstruction of the block. For this reason, we employ the Dual-Scale Sensing (DSS) matrix \cite{sankaranarayanan2012cs}. The key property of the DSS matrix is the ability to generate a low-resolution preview of the image with low complexity, while at the same time preserving the properties that enable CS reconstruction. A DSS matrix is obtained as
$$
\Ph_\mathsf{DSS} = \mat{H}\mat{D}+\mat{F}~,
$$
where $\mat{H}$ is an $M\times M$ Hadamard matrix, having $\pm 1$ elements such that $\mat{H}\trasp{\mat{H}} = M\mat{I}_M$, $\mat{D}$ is an $N-\mathrm{to}-M$ downsampling operator and $\mat{F}$ contains a random pattern such that $\mat{F}\trasp{\mat{D}} = \mat{0}$. This construction preserves matrix properties from the CS point of view, thanks to the contribution of matrix $\mat{F}$, and also allows to obtain a fast preview of the image at low resolution, since it minimizes the error between the downsampled version of the signal and the computed preview. The preview is generated as 
\begin{equation}\label{eq:preview}
\x_{\mathsf{P},i} = \mat{H}^{-1}\y_i^,
\end{equation}
where $\x_{\mathsf{P},i}$ contains the rows of the preview stacked on top of each other. This operation is fast since $\mat{H}$ is a Hadamard matrix and the inversion can be implemented by the fast Walsh-Hadamard transform algorithm \cite{1674569}. Moreover, the use of a Hadamard matrix imposes some constraints on the number of measurements $M$. In particular, $M$ must be a power of 2 and a perfect square, since the preview will be of size  $\sqrt{M} \times \sqrt{M}$ pixels.

\subsection{Image reconstruction}\label{sec:image_reconstruction}

\begin{figure}
\centering
\includegraphics[width=0.7\columnwidth]{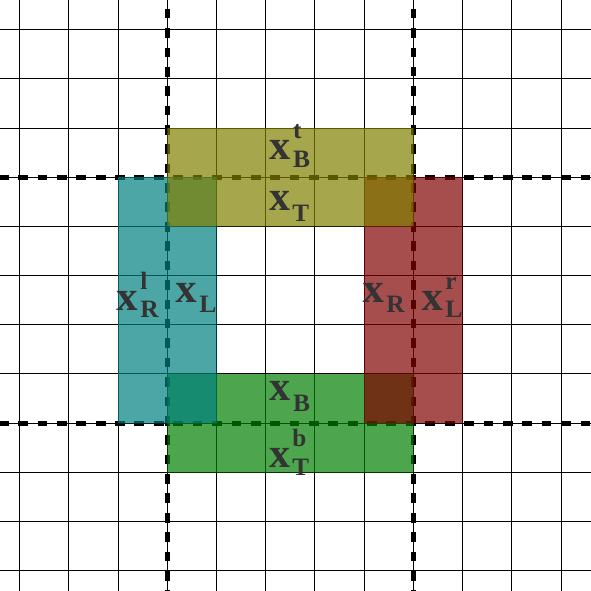}
\caption{Similarity constraints between adjacent block borders}
\vspace*{-5mm}
\label{fig:additional_constraints}
\end{figure}

SC-BCS aims at maximizing the reconstruction quality by minimizing the visual effect of blocking artifacts. In particular, this is achieved by adding some additional smoothness constraints among neighbouring blocks to the reconstruction problem of \eqref{eq:CS_TV_recovery}. In particular (see Fig.~\ref{fig:additional_constraints} for reference), reconstruction of current block \X\ is obtained as the solution of the following problem
\begin{equation} \label{eq:improved_reconstruction}
\widehat{\X}=\arg\min_{\X}\mathrm{TV}(\X)\ \quad \text{s.t.}\quad 
\begin{cases}
\Ph\cdot\vect{\X} &= \y  \\
\ltwonorm{\x_\mathsf{T}-\x^t_\mathsf{B}} &< \varepsilon_\mathsf{T}  \\
\ltwonorm{\x_\mathsf{B}-\x^b_\mathsf{T}} &< \varepsilon_\mathsf{B} \\
\ltwonorm{\x_\mathsf{L}-\x^l_\mathsf{R}} &< \varepsilon_\mathsf{L}  \\
\ltwonorm{\x_\mathsf{R}-\x^r_\mathsf{L}} &< \varepsilon_\mathsf{R}~,
\end{cases}
\end{equation}
where $\x_\mathsf{T}$ is the top row of current block, $\x_\mathsf{B}$ is the bottom row of current block, $\x_\mathsf{L}$ is the leftmost column of current block, $\x_\mathsf{R}$ is the rightmost column of current block, $\x^t_\mathsf{B}$ is the bottom row of the block on top of current one, $\x^b_\mathsf{T}$ is the top row of the block at the bottom of current one, $\x^l_\mathsf{R}$ is the rightmost column of the block on the left of current one and $\x^r_\mathsf{L}$ is the leftmost column of the block on the right of current one. Parameters $ \varepsilon_\mathsf{T},  \varepsilon_\mathsf{B},  \varepsilon_\mathsf{L}\ \mathrm{and}\  \varepsilon_\mathsf{R}$ are automatically evaluated from the initial border estimation, as explained later. The principle behind \eqref{eq:improved_reconstruction} is to ``drive'' the minimization to promote the solutions enhancing the smoothness between adjacent blocks. The reconstruction problem \eqref{eq:improved_reconstruction} raises two implementation problems, to which we propose solution in the following.

The first problem is to obtain an estimate of the borders of all blocks to plug into equation \eqref{fig:image_acquisition} as $\x^t_\mathsf{B}$, $\x^b_\mathsf{T}$, $\x^l_\mathsf{R}$ and $\x^r_\mathsf{L}$ to obtain the reconstruction of each block. The answer to this problem is given by the fast preview, which is enabled by the use of the DSS matrix. Hence, before running equation \eqref{eq:improved_reconstruction} on each block, we first reconstruct a low resolution preview using \eqref{eq:preview}, then we interpolate it to obtain a preview of the block with the original resolution. Since the blocks are acquired in an overlapping fashion, we merge the previews of the blocks by properly averaging neighbouring borders to obtain a preview of the entire image at the original size.

The second problem is related to the fact that the measurement vector $\y_i$ contains measurements of pixels belonging to the current block as well as to the borders of its neighbouring blocks, while the problem in \eqref{eq:improved_reconstruction} aims to reconstruct the $i$-th block \emph{without} employing the pixels belonging to neighbouring blocks. Again, to solve this problem we use the previously estimated preview to subtract the contribution of pixels belonging to neighbouring blocks from $\y_i$
$$
\y_{i}^\mathsf{inside} = \y_i - \Ph_i^\mathsf{border}\x_{\mathsf{P},i}^\mathsf{border}~,
$$
where $\x_{\mathsf{P},i}^\mathsf{border}$ contains only the pixel of the preview belonging to neighbouring blocks and $\Ph_i^\mathsf{border}$ contains the corresponding columns of the sensing matrix.

\begin{figure}
\centering
\includegraphics[width=0.75\columnwidth]{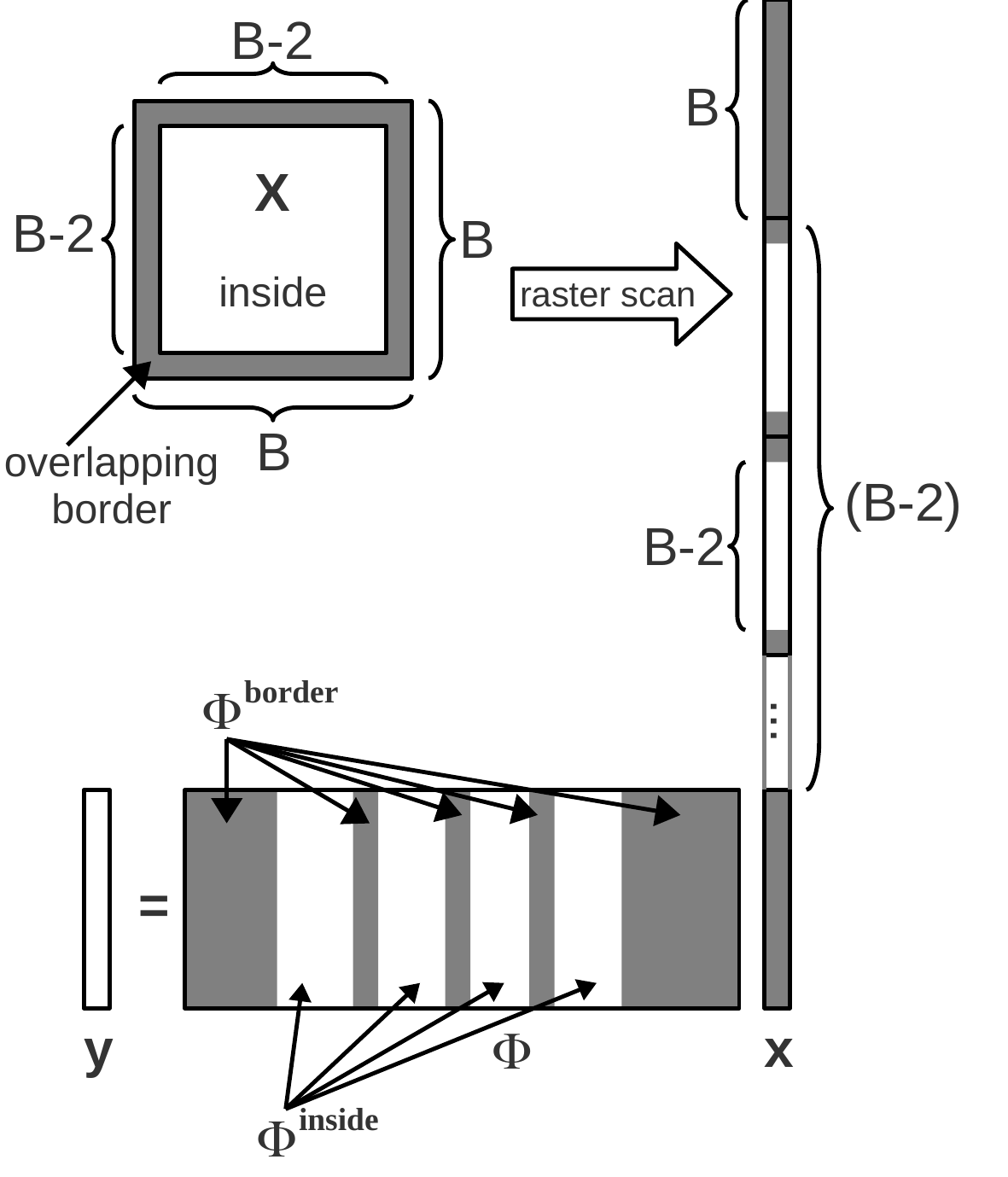}
\vspace{-3mm}
\caption{Block inside and border separation}
\vspace{-3mm}
\label{fig:separation}
\end{figure}

Finally, we run \eqref{eq:improved_reconstruction} for each block, using $\y = \y_{i}^\mathsf{inside}$ and $\Ph = \Ph_i^\mathsf{inside}$, where $\Ph_i^\mathsf{inside}$ contains the columns of the sensing matrix whose indexes correspond to the pixels of $i$-th block (excluding pixel belonging to neighbouring blocks due to overlapping). Fig.~\ref{fig:separation} visually explains the raster scanning of a block, highlighting the indexes belonging to block borders and their corresponding columns in the sensing matrix.

Finally, we propose to exploit the entire available preview (and not only its borders) as a predictor of the block, reconstructing with  \eqref{eq:improved_reconstruction} only the prediction error. This is justified by the assumption that, if the predictor is accurate enough, the prediction error will be more compressible than the signal itself. The reconstruction problem will then become
\begin{align}\label{eq:improved_reconstruction_predict}
\widehat{\X}&=\X_\mathsf{P}+\widehat{\mat{E}},\mathrm{where}\nonumber  \\
\widehat{\mat{E}}&=\arg\min_{\mat{E}}\mathrm{TV}(\mat{E})&\text{s.t.}\ 
\begin{cases}
\Ph\cdot\vect{\X_\mathsf{P} + \mat{E}} = \y  \\
\ltwonorm{\x_\mathsf{T}-\x^t_\mathsf{B}} < \varepsilon_\mathsf{T}  \\
\ltwonorm{\x_\mathsf{B}-\x^b_\mathsf{T}} < \varepsilon_\mathsf{B} \\
\ltwonorm{\x_\mathsf{L}-\x^l_\mathsf{R}} < \varepsilon_\mathsf{L}  \\
\ltwonorm{\x_\mathsf{R}-\x^r_\mathsf{L}} < \varepsilon_\mathsf{R}
\end{cases}
\end{align}
and $\X_\mathsf{P}$ corresponds to the rearrangement of the prediction $\x_\mathsf{P}$ in matrix form, $\x_\mathsf{T}$, $\x_\mathsf{B}$, $\x_\mathsf{L}$ and $\x_\mathsf{R}$ refer to matrix $\X_\mathsf{P} + \mat{E}$ and $\x^t_\mathsf{B}$, $\x^b_\mathsf{T}$, $\x^l_\mathsf{R}$, $\x^r_\mathsf{L}$, $ \varepsilon_\mathsf{T}, \varepsilon_\mathsf{B},  \varepsilon_\mathsf{L}\ \mathrm{and}\  \varepsilon_\mathsf{R}$ are evaluated as in \eqref{eq:improved_reconstruction}.

SC-BCS reconstruction is summarized in Algorithm~\ref{rec_algo}. We remark the fact that the most complex operation, consisting in the solution of \eqref{eq:improved_reconstruction_predict}, does not need any interaction among the blocks and hence can be run fully in parallel. On the other hand, other approaches such as BCS-SPL in \cite{mun2009block} require global operations needing synchronization among blocks.

\begin{algorithm}
\caption{Proposed algorithm}
\begin{algorithmic}
\For{$i = 1 \to N_{\mathsf{blocks}}$} 
\State $\x_{\mathsf{P},i} = \mat{H}^{-1}\y_i^,$ 
\State Interpolate $\x_{\mathsf{P},i}$ to the full block size
\EndFor
\State Merge the overlapping previews to obtain a preview of the original image $\x_{\mathsf{P}}$
\For{$i = 1 \to N_{\mathsf{blocks}}$} 
\State $\y_{i}^\mathsf{inside} = \y_i - \Ph_i^\mathsf{border}\x_{\mathsf{P},i}^\mathsf{border}$
\State Reconstruct the block using \eqref{eq:improved_reconstruction_predict}.
\EndFor
\end{algorithmic}
\label{rec_algo}
\end{algorithm}

\vspace*{-1mm}
\section{Numerical Results}

\begin{table*}
\caption{Obtained PSNR (dB) / SSIM index for various settings}
\centering
\begin{tabular}{c c | c c c c c c c}
Scheme & $M$ per block & \emph{Lena} & \emph{Goldhill} & \emph{Peppers} & \emph{Barbara} & \emph{Mandrill} & \emph{Boat} & \emph{Couple}\\
\hline
Independent & 73 & 25.79 / 0.800 & 25.56 / 0.737 & 25.16 / 0.808 & 21.95 / 0.691 & 19.98 / 0.553 & 23.49 / 0.713 & 23.46 / 0.693 \\
SC-BCS & 64 & \textbf{28.49} / \textbf{0.893} & \textbf{27.32} / \textbf{0.837} & \textbf{27.59} / \textbf{0.898} & \textbf{23.01} / \textbf{0.768} & \textbf{20.82} / \textbf{0.662} & \textbf{25.27} / \textbf{0.815} & \textbf{25.07} / \textbf{0.803}\\
BCS-SPL DDWT & 73 & 26.54 / 0.847 & 26.14 / 0.761 & 27.54 / 0.880 & 22.10 / 0.731 & 20.25 / 0.551 & 24.09 / 0.749 & 24.00 / 0.724\\
\hline
SC-BCS Baseline & 64 & 28.31 / 0.889 & 27.30 / 0.832 & 27.56 / 0.899 & 23.13 / 0.773 & 20.94 / 0.658 & 25.20 / 0.810 & 25.07 / 0.798\\
SC-BCS Genie & 64 & 28.79 / 0.903 & 27.64 / 0.848 & 28.54 / 0.912 & 23.38 / 0.789 & 21.04 / 0.678 & 25.65 / 0.834 & 25.52 / 0.822\\
\hline
\hline
Independent & 289 & 31.08 / 0.939 & 29.61 / 0.911 & 30.14 / 0.928 & 24.56 / 0.846 & 22.62 / 0.813 & 28.48 / 0.899 & 27.72 / 0.900 \\
SC-BCS & 256 & 32.21 / 0.950 & 29.93 / \textbf{0.919} & 31.29 / 0.939 & 24.06 / 0.822 & \textbf{22.81} / \textbf{0.809} & 28.34 / 0.906 & \textbf{28.34} / \textbf{0.909} \\
BCS-SPL DDWT & 289 & \textbf{33.21} / \textbf{0.960} & \textbf{30.13} / 0.917 & \textbf{33.49} / \textbf{0.961} & \textbf{25.38} / \textbf{0.875} & 22.71 / 0.798 & \textbf{29.24} / \textbf{0.916} & 28.33 / 0.906\\
\hline
SC-BCS Baseline & 256 & 31.63 / 0.946 & 29.99 / 0.918 & 30.93 / 0.940 & 24.58 / 0.842 & 22.48 / 0.815 & 28.19 / 0.905 & 28.11 / 0.902 \\
SC-BCS Genie & 256 & 31.78 / 0.957 & 30.22 / 0.933 & 31.14 / 0.949 & 24.82 / 0.870 & 22.85 / 0.839 & 28.47 / 0.922 & 28.40 / 0.925 \\
\hline
\hline
\end{tabular}
\label{tab:results}
\vspace*{-5mm}
\end{table*}

\begin{figure}
\centering
\subfigure[]{\includegraphics[width=0.7\columnwidth]{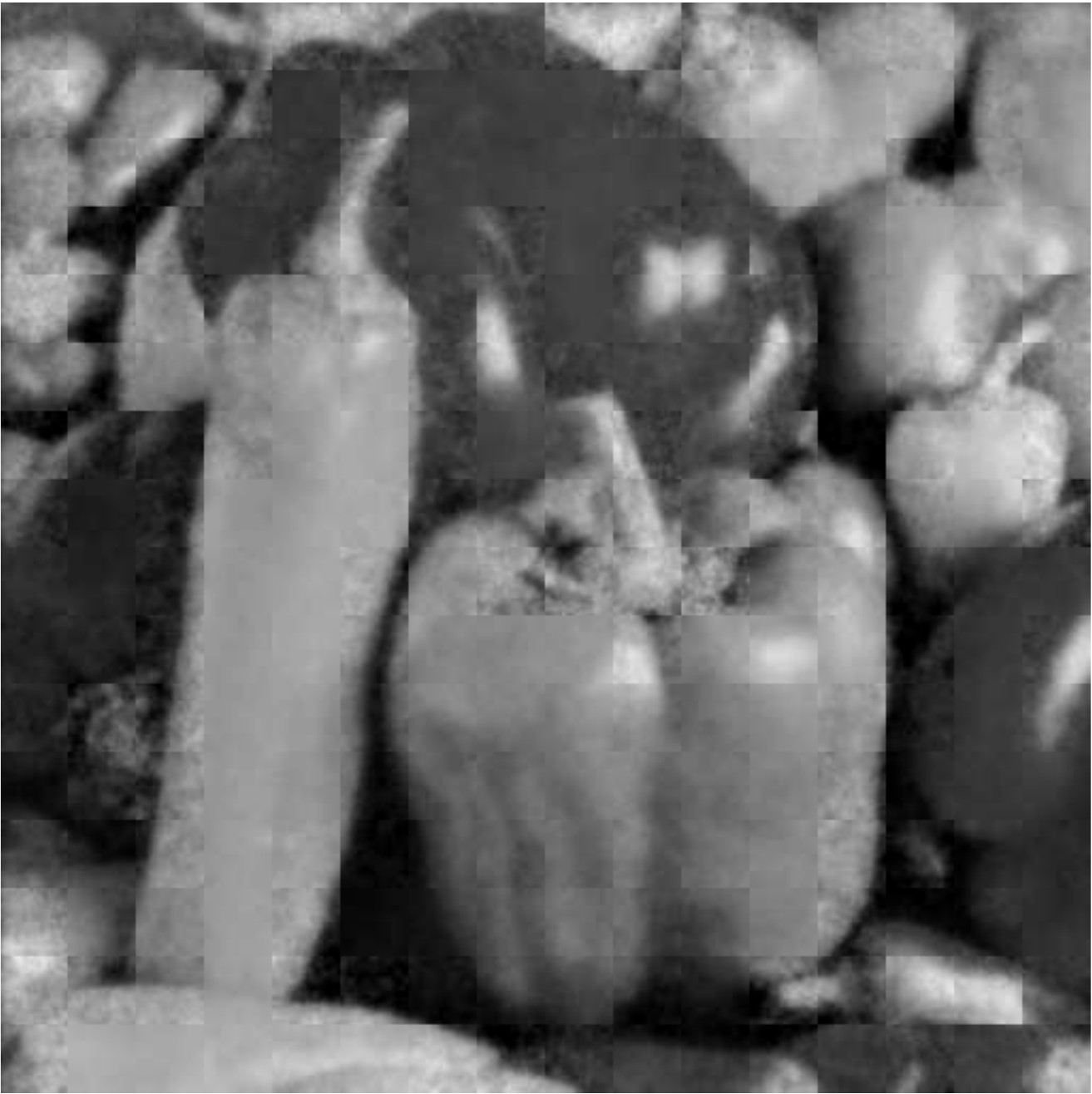}\label{fig:peppers_standard_gauss_M73}}\quad
\subfigure[]{\includegraphics[width=0.7\columnwidth]{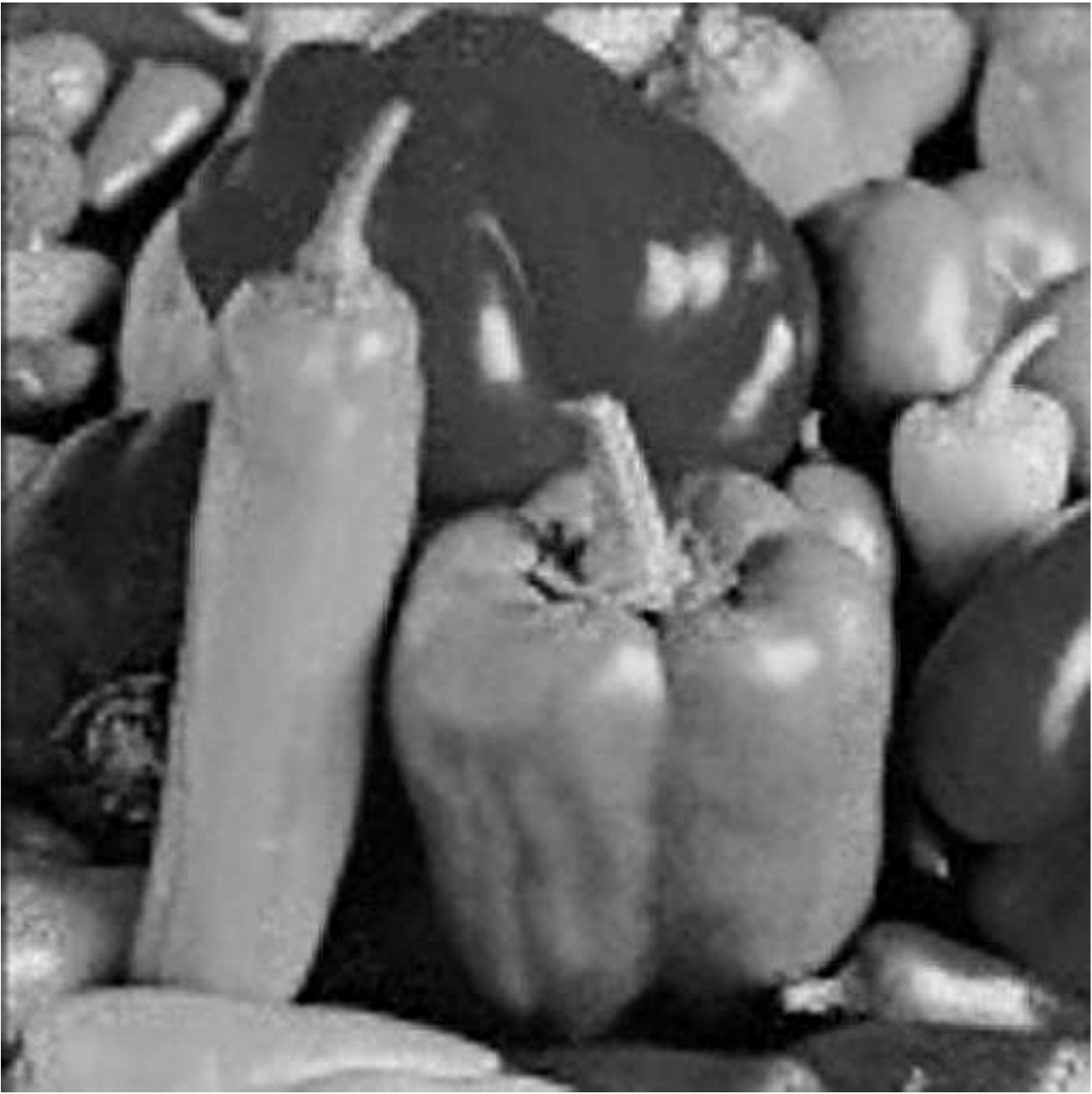}\label{fig:peppers_proposed_M64}}
\vspace*{-1mm}
\caption{\emph{Peppers}, top down: blockwise independent reconstruction with $M=73$ measurements per block, SC-BCS with $M=64$ measurements per block}
\vspace*{-4mm}
\end{figure}

SC-BCS has been tested using various grayscale test images of size $512 \times 512$. Each image is divided into blocks, whose interior has size $30 \times 30$ plus 1 extra pixel on each side. Hence, this corresponds to $B=32$ and each block is then composed by $N=B^2=1024$ pixels. We test the proposed algorithm with  $M=64$ and $M=256$ measurements per block, corresponding to compression rates of $93\%$ and $72\%$ compression. This corresponds to splitting the image in $17^2=289$ overlapping blocks. We compare with two systems performing non overlapped acquisition using a Gaussian random sensing matrix, where each element of the matrix is distributed as $\N(0,1/M)$. The first one reconstructs each block independently using \eqref{eq:CS_TV_recovery}, while the second one is the BCS-SPL-DDWT scheme\footnote{For BCS-SPL-DDWT, the software provided by the authors at \texttt{http://www.ece.msstate.edu/$\sim$fowler/BCSSPL/} has been used} \cite{mun2009block}, where independent reconstruction is followed by a \emph{global} iterative image processing stage employing Wiener filtering and thresholding of the coefficients of the image projected on a directional transform. For a fair comparison, since the number of non overlapping $32 \times 32$ blocks is lower ($16^2=256$), for these systems we will take a greater number of measurements per block ($M=73$ and $M=289$, respectively) so that the total number of measurements is the same as in the overlapped case.

Table~\ref{tab:results} summarizes the results in terms of PSNR and SSIM index \cite{wang2004image}.  SC-BCS Baseline refers to the scheme not using the preview as a predictor \eqref{eq:improved_reconstruction}, while SC-BCS Genie refers to an ideal scheme where reconstruction \eqref{eq:improved_reconstruction} is performed knowing perfectly block borders. Results show that the proposed idea leads to significant gains. It can be noticed that when $M=64$ SC-BCS leads to significant gains with respect to the independent case, proving the validity of the approach. Moreover, it outperforms BCS-SPL-DDWT with all images. On the other hand, when $M=256$ the gains with respect to independent reconstruction are lower and BCS-SPL-DDWT has the same or better performance. The reason is that when $M$ is large, even independent reconstruction leads to good estimates, hence the contribution of additional smoothness constraints is less significant. Performance obtained by SC-BCS Baseline and SC-BCS Genie is reported to show that imperfect knowledge of block borders implies a loss of only $0.1\div0.3$~dB with respect to the ideal case where borders are perfectly known.

\begin{figure*}
\centering
\vspace*{-1mm}
\subfigure[]{\includegraphics[width=5.4cm]{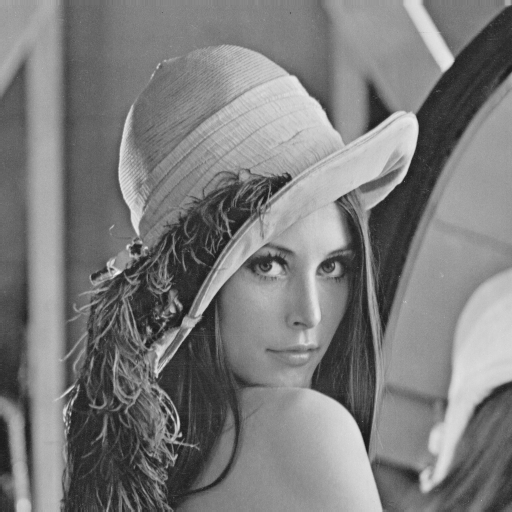}\label{fig:lena_orig}}\quad
\subfigure[]{\includegraphics[width=5.4cm]{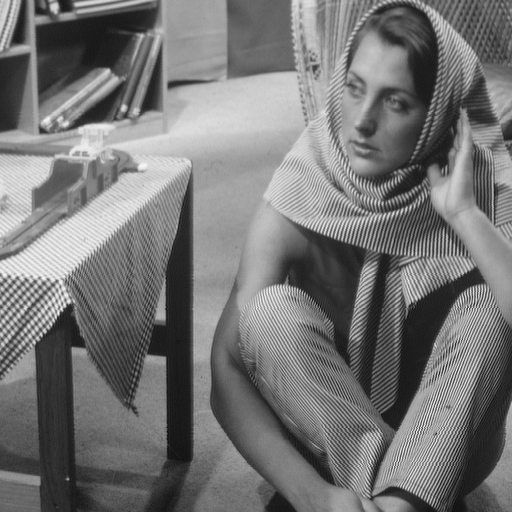}\label{fig:barbara_orig}}\quad
\subfigure[]{\includegraphics[width=5.4cm]{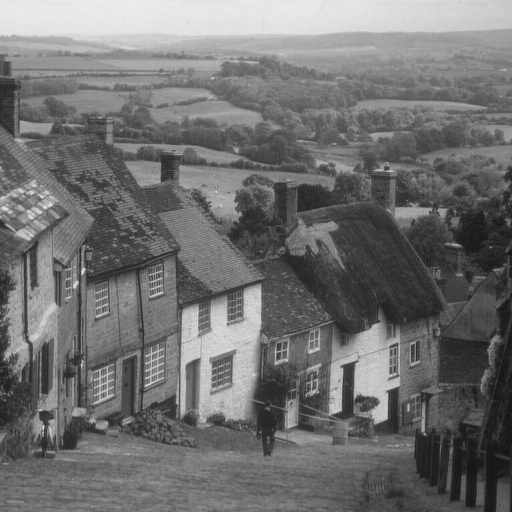}\label{fig:goldhill_orig}}\quad
\subfigure{\includegraphics[width=5.4cm]{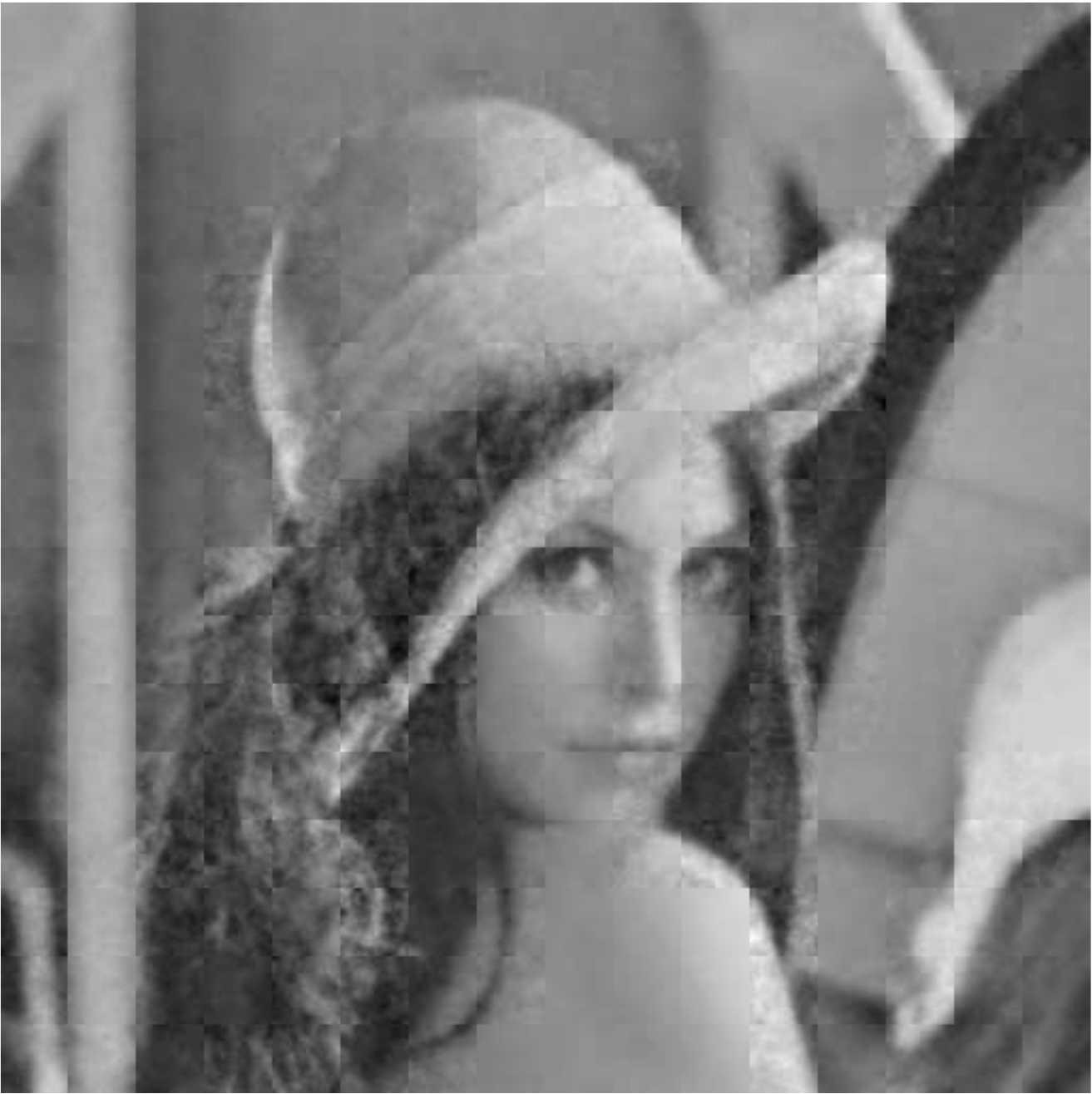}\label{fig:lena_standard_gauss_M73}}\quad
\subfigure{\includegraphics[width=5.4cm]{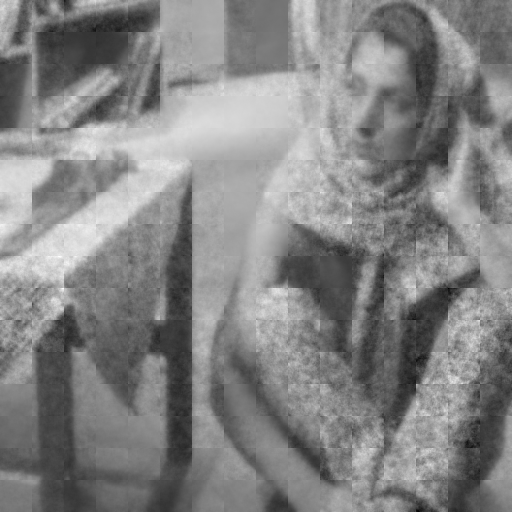}\label{fig:barbara_standard_gauss_M73}}\quad
\subfigure{\includegraphics[width=5.4cm]{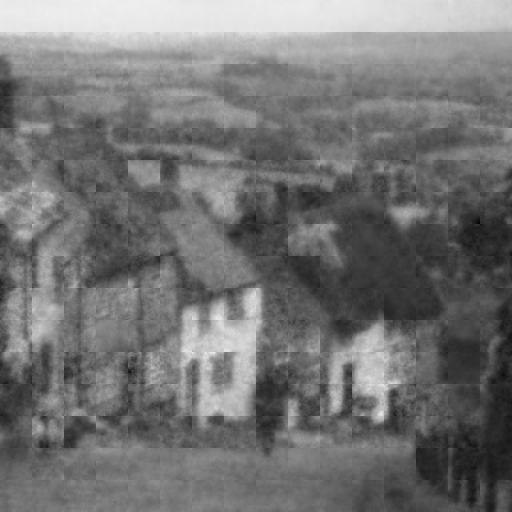}\label{fig:goldhill_standard_gauss_M73}}\quad
\subfigure{\includegraphics[width=5.4cm]{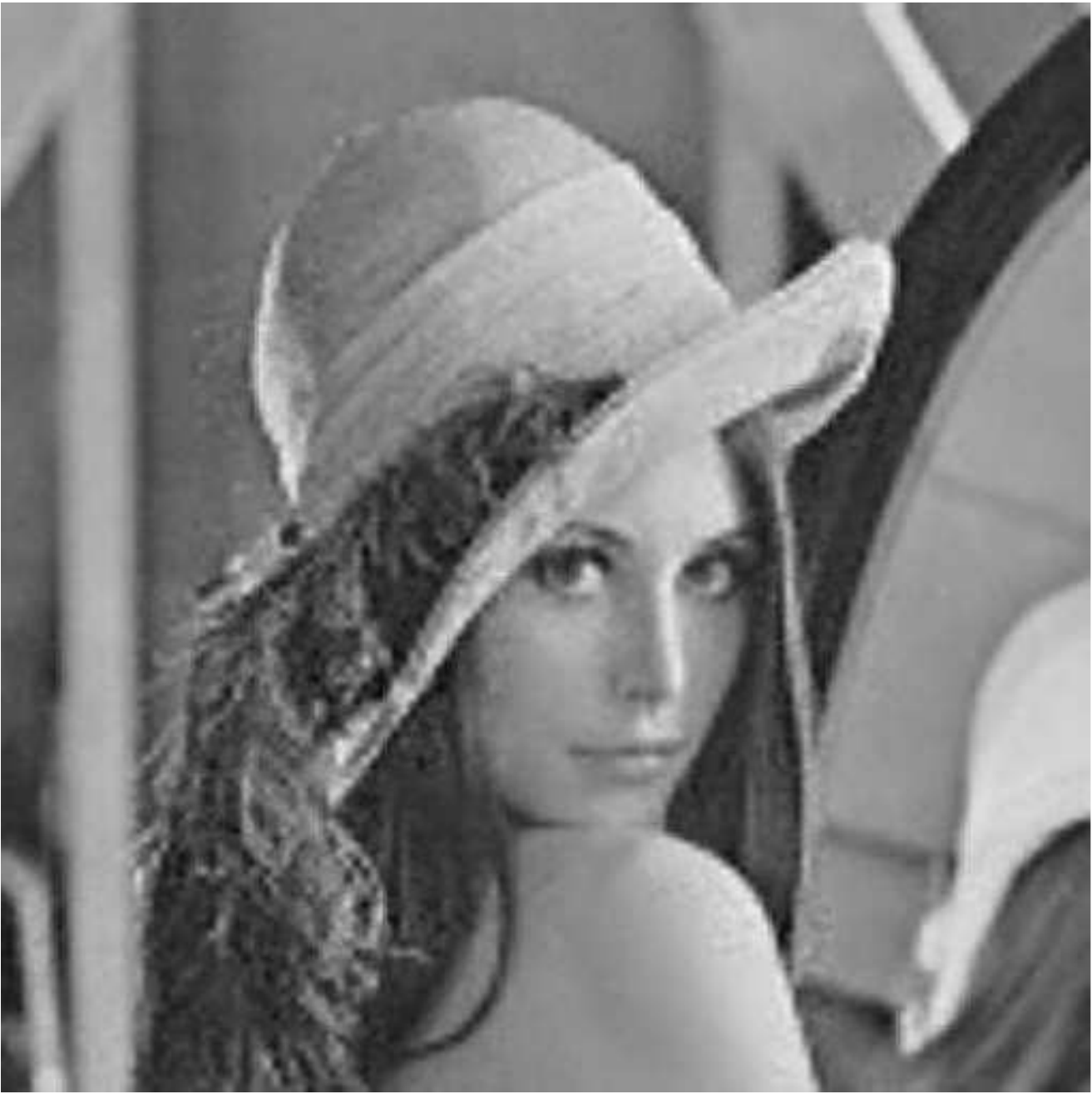}\label{fig:lena_proposed_M64}}\quad
\subfigure{\includegraphics[width=5.4cm]{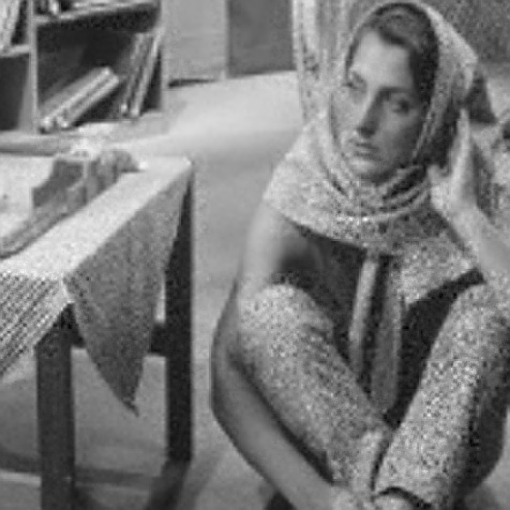}\label{fig:barbara_proposed_M64}}\quad
\subfigure{\includegraphics[width=5.4cm]{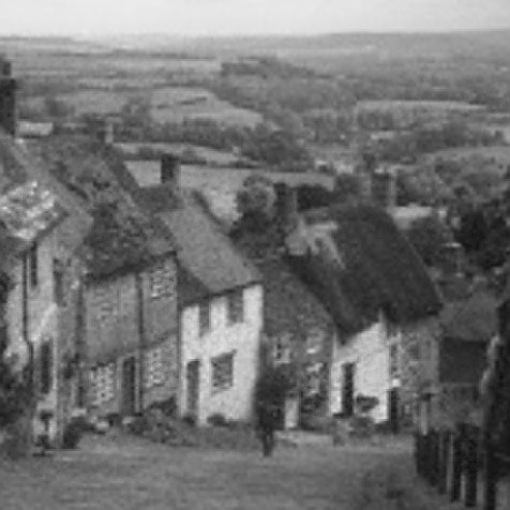}\label{fig:goldhill_proposed_M64}}\quad
\subfigure{\includegraphics[width=5.4cm]{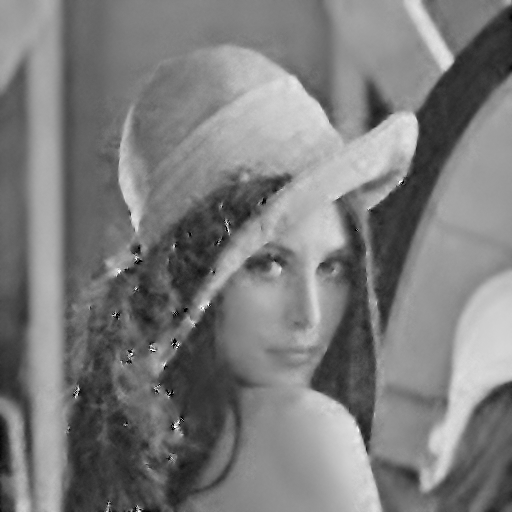}\label{fig:lena_SPL_M73}}\quad
\subfigure{\includegraphics[width=5.4cm]{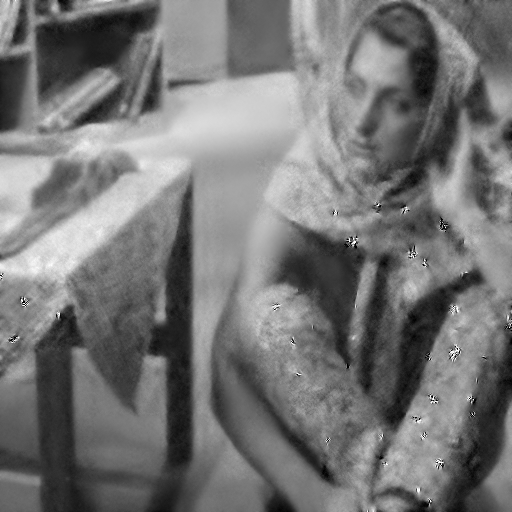}\label{fig:barbara_SPL_M73}}\quad
\subfigure{\includegraphics[width=5.4cm]{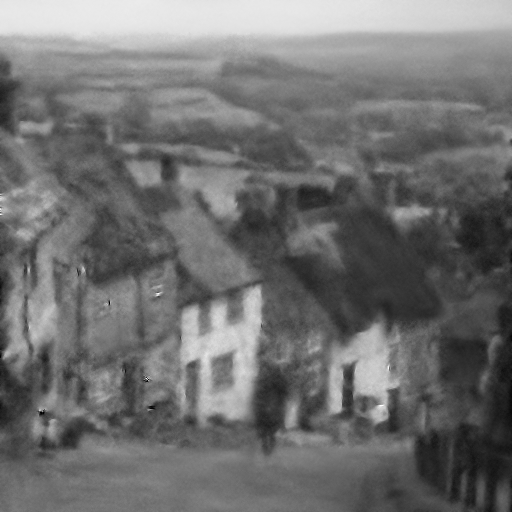}\label{fig:goldhill_SPL_M73}}\quad
\caption{(a) \emph{Lena}, (b) \emph{Barbara}, (c) \emph{Goldhill}. Top down: original, blockwise independent reconstruction with $M=73$ measurements per block, SC-BCS (proposed) with $M=64$ measurements per block, BCS-SPL-DDWT with $M=73$ measurements per block}
\label{fig:summary}
\end{figure*}

We conclude by showing some reconstructed images to assess the visual quality of the reconstruction. In fact we stress that we are addressing a visual quality problem and an increase in PSNR does not necessarily imply a reduction in the blockiness of the image. We report only the case with $M=64$, where the differences among various reconstruction schemes are more easily noticeable. However, we remark that less blocky artifacts can be noticed in the $M=256$ case, as well. Fig.~\ref{fig:peppers_standard_gauss_M73} shows the results of the independent blockwise reconstruction while Fig.~\ref{fig:peppers_proposed_M64} shows the reconstructed image provided by SC-BCS. It can be clearly noticed that independent reconstruction yields the poorest visual reconstruction quality, while our proposed scheme shows fewer blocking artifacts, with an acceptable visual quality at a compression rate of $93\%$. Figs.~\ref{fig:summary} shows a visual comparison among independent recovery, SC-BCS and BCS-SPL-DDWT. It can be noticed that SC-BCS has the best visual quality performance, with less blocky artifacts with respect to the independent case. On the other hand, when $M=73$ BCS-SPL-DDWT reconstruction suffers from localized clusters of out-of-scale pixel values, which significantly deteriorate the quality of the image, and tends to flatten image details.

\section{Conclusions}
\vspace*{-0.5mm}

In this paper we proposed a method to address a visual quality issue of block-based compressed sensing. Blocking artifacts due to independent reconstruction of the blocks can severely affect the quality of the final image. We showed that imposing additional constraints in the reconstruction problem allows to exploit prior information on the smoothness of the image across block borders. We solved a dependency problem in which the constraints depend on the reconstructions themselves by using a fast preview of each block and, by combining the overlapping parts of the previews, it was possible to enforce a smooth reconstruction. Finally we showed that the already available preview can be used as an additional signal model besides sparsity. This information was used as a prediction of the final recovered image to improve the quality of the reconstruction. 


\end{document}